\documentclass[twocolumn]{cinc}
\usepackage[utf8]{inputenc}
\usepackage[T1]{fontenc}
\usepackage{graphicx}
\usepackage{multirow}
\usepackage{hhline}
\usepackage{ctable}
\usepackage[noend]{algpseudocode}
\usepackage{algorithm}
\usepackage{amsmath}
\usepackage{url}
\usepackage{hyperref}

\hypersetup{colorlinks=false,pdfborder={0 0 0},
bookmarksopen,bookmarksdepth=3}

\begin{document}

\title{Arrhythmia Classification from the Abductive Interpretation of Short 
Single-Lead ECG Records}


\author{Tomás Teijeiro*, Constantino A. García, Daniel Castro and Paulo Félix\\
\ \\ 
 Centro Singular de Investigación en Tecnoloxías da Información (CITIUS),
University of Santiago de Compostela, Santiago de Compostela, Spain }

\maketitle

\begin{abstract}
\vspace{-0.75em}
In this work we propose a new method for the rhythm classification of short 
single-lead ECG records, using a set of high-level and clinically meaningful 
features provided by the abductive interpretation of the records. These 
features include morphological and rhythm-related features that are used to 
build two classifiers: one that evaluates the record globally, using aggregated 
values for each feature; and another one that evaluates the record as a 
sequence, using a Recurrent Neural Network fed with the individual features for 
each detected heartbeat. The two classifiers are finally combined using the 
stacking technique, providing an answer by means of four target classes: 
Normal sinus rhythm (\textbf{N}), Atrial fibrillation (\textbf{A}), 
Other anomaly (\textbf{O}) and Noisy (\textbf{\textasciitilde}). The approach 
has been validated against the 2017 Physionet/CinC Challenge dataset, obtaining 
a final score of 0.83 and ranking first in the competition.
\end{abstract}

\section{Introduction}
The potential of Artificial Intelligence and machine learning techniques to 
improve the early detection of cardiac diseases using low-cost ECG tests is 
still largely untapped. The 2017 Physionet/Computing in Cardiology challenge 
defies the scientific community to propose solutions to the automatic detection 
of Atrial Fibrillation from short single lead ECG signals~\cite{Clifford17}. The 
challenge is posed as a classical machine learning problem: A labeled training 
set is provided, and the proposals are evaluated against a hidden test set of 
records. However, even if the only metric for the final ranking is the accuracy 
of the proposed models, a number of additional properties should be considered 
for the final adoption of each proposal in the clinical practice. Here, we 
emphasize on the interpretability of the automatic detection of Atrial 
Fibrillation, a major concern to ensure trust by the care 
staff~\cite{Caruana15}. 

In this sense, our proposal is based on a high-level description of the target 
signal by means of the same features used by cardiologists in ECG analysis. This 
description is generated with a pure knowledge-based approach, using an 
abductive framework for time series interpretation~\cite{Teijeiro16} that looks 
for the set of explanatory hypotheses that best account for the observed 
evidence. Only after this description has been built, machine learning methods 
were used to make up for the lack of the expert criteria applied in the labeling 
of the training set, and to alleviate the effect of possible errors in the 
interpretation process.

\section{Methods}
\label{sec:methods}

\begin{figure*}[t]
\centering
\includegraphics[width=\textwidth]{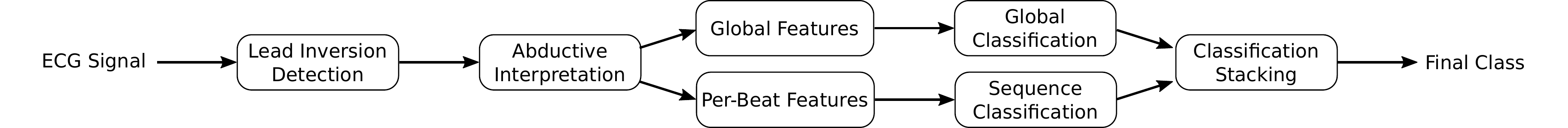}
\caption{Classification algorithm steps.}
\label{fig:arch}
\end{figure*}

The global architecture of the proposal is depicted in Figure~\ref{fig:arch}, 
and the processing stages are explained in the following subsections.

\subsection{Preprocessing}
The preprocessing stage aims at improving the quality of the data to be 
interpreted in the following stages, and involves two different tasks: 

\begin{enumerate}
 \item[2.1.1.] \textit{Data relabeling:} The labeling of the training set was 
performed by a single expert in a single pass, and as a consequence some 
inconsistencies appear in the classification criteria. Thus, a thorough manual 
relabeling was carried out, but trying to be conservative and guided by 
pre-\linebreak liminary classification results. We focused on records 
classified as \textbf{N} but showing what we consider clear anomalies. A total 
number of 197 out of 8528 records were relabeled.
 \item[2.1.2.] \textit{Lead inversion detection:} A number of records in the 
training set were found to be inverted, probably due to electrode misplacement. 
Inverted records are more likely to be classified as abnormal due to the 
presence of infrequent QRS and T wave morphologies, as well as to the greater 
difficulty to identify P waves. The inverted records were first identified 
manually, and then a simple logistic regression classifier was trained 
considering 14 features obtained from the raw signal and a tentative delineation 
of the P wave, QRS complex and T wave of every heartbeat detected by the 
\texttt{gqrs} application from the Physionet library~\cite{Goldberger00}. This 
delineation was performed using the \textit{Construe} 
algorithm~\cite{Teijeiro16}, limiting the interpretation to the conduction 
level, that is, avoiding the rhythm interpretation. 
\end{enumerate}

\subsection{Abductive interpretation}
\begin{figure*}[t]
\centering
\includegraphics[width=\textwidth]{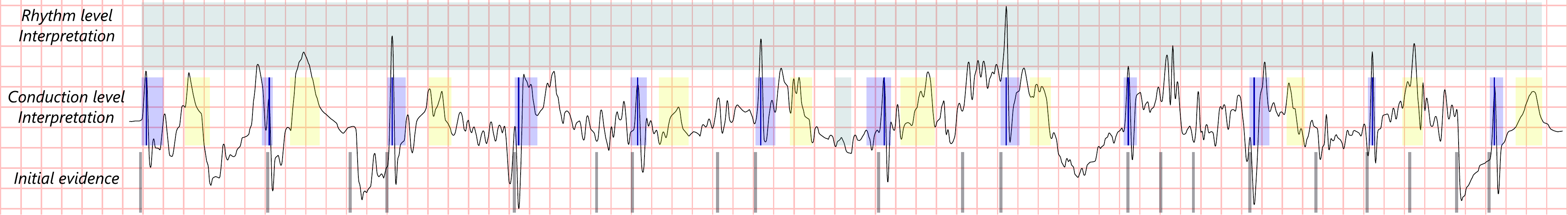}
\caption{How the abductive interpretation can fix errors in the initial 
evidence. {\footnotesize [Source: First 10 seconds of the A02080 record. Grey: 
Original \texttt{gqrs} annotations. Blue: QRS observations. Yellow: T wave 
observations. Green: P wave observation and Normal rhythm hypothesis.]}}
\label{fig:error_fix}
\end{figure*}

The abductive interpretation of the ECG signal is the most significant stage in 
the proposed approach. Its objective is to characterize the physiological 
processes underlying the signal behavior, building a description of the observed 
phenomena in multiple abstraction levels. This responsibility lies with the 
\textit{Construe} algorithm, which applies a non-monotonic reasoning scheme to 
find the set of hypotheses that best explain the observed evidence, by means of 
a domain-specific knowledge base composed of a set of \textit{observables} and a 
set of \textit{abstraction grammars}. The knowledge base is the same used 
in~\cite{Teijeiro16}, that allows to explain the ECG at the conduction and 
rhythm abstraction levels, thus providing the same features used by 
cardiologists in ECG analysis. The initial evidence is the set of waves 
identified in the wave delineation step, that are abstracted by a set of rhythm 
patterns to describe the full signal as a sequence of cardiac rhythms, including 
normal sinus rhythms, bradycardias, tachycardias, atrial fibrillation episodes, 
etc. The non-monotonic nature of the interpretation process allows us to modify 
the initial set of evidence, by discarding heartbeats that cannot be abstracted 
by any rhythm pattern, or by looking for missed beats that are predicted by the 
pattern selected as the best explanatory hypothesis for a signal fragment. This 
ability to correct the initial evidence is the main strength of our proposal, 
since it discards many false anomalies generated by the presence of noise and 
artifacts in the signal. Figure~\ref{fig:error_fix} shows an example of a noisy 
signal in which the \texttt{gqrs} application detects many false positive beats, 
that are removed or modified in the final interpretation that concludes with a 
single normal rhythm hypothesis that explains the full fragment. As we can also 
see in the Figure, the result of the interpretation stage is a sequence of P 
waves, QRS complexes and T waves observations, as well as a sequence of cardiac 
rhythms abstracting all those waves.

\subsection{Global feature extraction}
Considering that each ECG record has to be classified globally, providing a 
single label for the entire signal duration, after the interpretation stage a 
set of features are calculated trying to summarize the information provided by 
\textit{Construe}. A total number of 79 features are calculated, that are 
comprehensively described in the published software documentation. The feature 
set is divided into three main groups:
\begin{itemize}
 \item \textit{Rhythm features:} This includes statistical measures on the RR 
sequence, such as the limits, median or median absolute deviation; heart rate 
variability features such as the PNN5, PNN10, PNN50 and PNN100 
measures~\cite{Mietus02}; and information about the rhythm interpretation, such 
as the median duration of each rhythm hypothesis.
 \item \textit{Morphological features:} This includes information about the 
duration, amplitude and frequency spectrum of the observations in the 
conduction abstraction level, including P and T waves, QRS complexes, PR and QT 
intervals, and the TP segments.
 \item \textit{Signal quality features:} Their purpose is to assess the 
importance of the morphological features showing conduction anomalies, such as 
wide QRS complexes or long PR intervals. They are based on the sum of the 
absolute differences of the signal, which we refer to as \textit{profile}. Some 
of the profiled areas of the signal are the baseline segments and the P wave 
area before each heartbeat (taking a constant window of 250 ms).
\end{itemize}

\subsection{Global classification}
If a precise definition of the expert knowledge leading to the labeling of the 
training set were available, then the final classification could be directly 
developed with a basic rule-based system operating on the features extracted 
from the abductive interpretation stage, and the accuracy of the system would 
depend mainly on the accuracy of the interpretations. However, the challenge 
does not publish any guidelines for the classification, specially for the 
\textbf{O} class. Therefore, an automatic classifier was trained with two 
objectives: 1) To reveal the criteria leading to the training set labeling; and 
2) to make the classification more accurate by learning possible mistakes of the 
abductive interpretation.

The classification method selected for this stage was the Tree Gradient Boosting 
algorithm, and particularly the XGBoost implementation~\cite{Chen16}, which 
showed a high performance and a certain level of interpretability through the 
importance given to the classification features. The optimization of the 
hyperparameters was performed using exhaustive grid search and 8-fold 
cross-validation, leading to the following values: Maximum tree depth: 6, 
Learning rate: 0.2, Gamma: 1.0, Column subsample by tree: 0.9, Min. child 
weight: 20, Subsample: 0.8, and Number of boosting rounds: 60.

With respect to the first objective, we were able to formalize a number of 
specific anomalies that lead to classify a record as \textbf{O}. This 
identification helped to optimize the training set by defining more specific 
features to be calculated from the interpretation results. Some of the 
identified anomalies sharing this class were:

\begin{itemize}
 \item Tachycardia (Mean heart rate over 100 bpm).
 \item Bradycardia (Mean heart rate under 50 bpm).
 \item Wide QRS complex (Longer than 110 milliseconds).
 \item Presence of ventricular or fusion beats.
 \item Presence of at least one extrasystole.
 \item Long PR interval (Longer than 210 milliseconds).
 \item Ventricular tachycardia.
 \item Atrial flutter.
\end{itemize}

For some of these anomalies the classification in the training set seems a bit 
inconsistent, since examples can be found in several classes. For example, there 
are various records labeled as normal with PR interval longer than 210 
milliseconds, as long as examples of records labeled as atrial fibrillation 
showing an atrial flutter pattern.

Regarding the second objective, even after discovering some of the expert 
criteria distinguishing the target classes a rule-based system was not still 
competitive against automatic learned models. From our point of view this shows 
that the XGBoost classifier is able to improve the results of the 
interpretation alone.

\subsection{Per-beat feature extraction\label{sec:per_beat}}

Some of the conditions leading to a certain classification may not be present 
for the entire duration of a record, so the global features are not the best 
option to characterize episodic events of abnormalities. For example, a normal 
record with a single ectopic ventricular beat that does not break the rhythm is 
quite difficult to classify as abnormal by the global classifier. For this 
reason, some of the features calculated from the abductive interpretation are 
disaggregated to the individual heartbeat scope, such as the morphology, 
duration and amplitude of the P wave, the QRS complex and the T wave. Also the 
RR interval and the RR variation before and after each beat is included, as 
long as the profile of the P wave area. A sequence classification approach is 
then used to learn characteristic temporal patterns of each target class.


\subsection{Sequence classification}
\begin{figure*}[t]
\centering
\includegraphics[width=\textwidth]{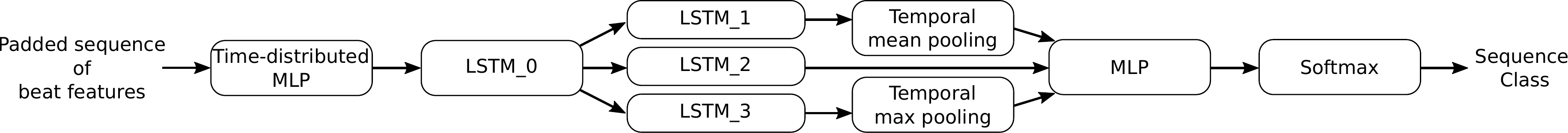}
\caption{The neural network architecture.}
\label{fig:nnet_arch}
\vspace{-0.25cm}
\end{figure*}

In the proposed approach, sequence classification relies on Recurrent Neural 
Networks (RNNs), a family of neural networks specialized for recognizing 
sequences of values. Among the different RNN implementations, we focused on Long 
Short Term Memory networks  (LSTMs) \cite{hochreiter1997long}, since they are 
capable of remembering information for long periods of time through the use of a 
cell state. Furthermore, they are able to avoid vanishing and exploding 
gradients when doing backpropagation through time. The architecture of the 
neural net is shown in Figure \ref{fig:nnet_arch}. The time-distributed 
Multilayer Perceptron (MLP) preprocesses the features described in Section 
\ref{sec:per_beat} to transform the data into a space with easier temporal 
dynamics. The number of hidden units of the MLP was 256, and the dimension of 
the output space 128. A Rectified Linear Unit (ReLU) was used as activation 
function. The \textit{LSTM\_0} layer preprocesses the resulting sequence of 
transformed features and returns a new sequence, which is subsequently used by 
the other LSTMs. The \textit{LSTM\_2} layer just returns the final state of the 
network, whereas \textit{LSTM\_1} and \textit{LSTM\_3} return new transformed 
sequences. The pooling layers after \textit{LSTM\_1} and \textit{LSTM\_3} remove 
the temporal dimension by computing the temporal mean and maximum of each 
feature of the sequences, respectively. All the LSTMs used 128 units. Another 
MLP (with the same configuration of the time-distributed one) joins and 
transforms the outputs of each LSTM before a Softmax layer, which outputs a 
probability for each of the 4 classes. $L^2$-regularization was applied to all 
layers, using $10^{-4}$ as regularization strength. Finally, dropout was also 
used  to improve generalization by preventing feature co-adaptation 
\cite{srivastava2014dropout}.

The neural network was trained using the categorical cross-entropy as loss 
function, a batch size of 32, and Adam \cite{KingmaB14} as optimizer. 
Furthermore, 15\% of all the data was used as validation set to monitor the 
performance of the neural network. This permitted us to decrease the learning 
rate when the validation loss got stuck in a plateau and to avoid overfitting by 
using early stopping. The initial learning rate was set to $0.002$ and it was 
decreased by $\sqrt 2$ when the validation loss did not improve for at least 3 
epochs. Training was ended after 15 epochs without improvement.

\subsection{Classification stacking}
The XGBoost classifier based on global features and the RNN classifier based on 
the per-beat features were combined using the stacking technique. Stacking (also 
referred to as stacked generalization) involves training a new classification 
algorithm to combine the predictions of several classifiers 
\cite{wolpert1992stacked}. Usually, the stacked model achieves better 
performance than the individual models due to its ability to discern when each 
base model performs best and when it performs poorly. Prior to the application 
of stacking, the predictions of 3 RNNs were averaged to decrease the variance of 
the RNN classifier arising from the random initialization of the RNN weights and 
the random split between test and validation set. Averaging similar models also 
helps in reducing overfitting. Note that this averaging can be seen as a simple 
bagging method. The probabilities predicted by the XGBoost and the averaged RNNs 
are then combined through a Linear Discriminant Analysis (LDA) classifier, which 
acts as stacker. To avoid possible collinearity issues, only 3 probabilities 
from each model are used.

\section{Evaluation}
\label{sec:evaluation}
To evaluate the performance of the algorithm, we followed the challenge 
guidelines and metrics. The final score is assigned as the mean F1 measure of 
the \textbf{N}, \textbf{A}, and \textbf{O} classes. Table \ref{tab:folds} shows 
an example of the results that the proposed method is able to achieve using 
8-fold cross-validation. Note that the stacker usually achieves better scores 
than the base models and, furthermore, it has lower variance (not shown in the 
Table).

\begin{table}[!ht]
\small
\setlength\tabcolsep{2.75pt} 
\renewcommand{\arraystretch}{1.3}
\caption{Example of stratified 8-fold cross-validation.}
\label{tab:folds}
\centering
\begin{tabular}{ c r r r r r r r r r}
\hline \hline
 & \multicolumn{8}{c}{Fold Number}& \multirow{2}{*}{Mean}\\
\cline{2-9}
Method & 0&  1& 2& 3& 4& 5& 6& 7&\\
\hline
XGBoost&0.84&0.84&0.85&0.85&0.82&0.80&0.82&0.82&0.83\\
RNN&0.82&0.81&0.84&0.83&0.86&0.83&0.83&0.83&0.83\\\
LDA-stacker&0.85&0.84&0.86&0.86&0.85&0.83&0.84&0.85&0.85\\
\hline \hline
\end{tabular}
\vspace{-0.5cm}
\end{table} 

\section{Conclusions}     
\label{sec:conclusions}
This work proves that the combination of knowledge-based and learning-based 
approaches is effective to build classification systems that exploit  
sophisticated machine learning methods while maintaining a remarkable degree of 
interpretability by the use of high-level and meaningful features.

\section*{Acknowledgements}  
This work was supported by the Spanish Ministry of Economy and 
Competitiveness under project TIN2014-55183-R. Constantino A. Garc\'ia 
is also supported by the FPU Grant program from the Spanish Ministry of 
Education (MEC) (Ref. FPU14/02489).
\bibliographystyle{cinc}
\bibliography{bibliography}

\vspace{-1.65em}
\begin{correspondence}
Tomás Teijeiro Campo\\
Rúa de Jenaro de la Fuente Domínguez, S/N, CITIUS Building\\
15782 Santiago de Compostela, SPAIN\\
tomas.teijeiro@usc.es
\end{correspondence}

\end{document}